\documentclass[10pt,twocolumn,letterpaper]{article}

\usepackage[pagenumbers]{cvpr}      % To produce the REVIEW version
\usepackage[dvipsnames]{xcolor}
\usepackage[normalem]{ulem}

\definecolor{cvprblue}{rgb}{0.21,0.49,0.74}
\usepackage[pagebackref,breaklinks,colorlinks,citecolor=cvprblue]{hyperref}

\def\paperID{*****} % *** Enter the Paper ID here
\def\confName{CVPR}
\def\confYear{2024}

\title{
Does Data-Efficient Generalization Exacerbate Bias in Foundation Models?
}
\author{
Dilermando Queiroz\\
Universidade Federal de São Paulo\\
\and
Anderson Carlos\\
Instituto Federal de Goiás\\
\and
Maíra Fatoretto\\
Universidade Federal de São Paulo\\
\and
Luis Filipe Nakayama\\
Universidade Federal de São Paulo\\
\and
André Anjos\\
Idiap Research Institute\\
\and
Lilian Berton\\
Universidade Federal de São Paulo\\
}
\begin{document}
\maketitle
\begin{abstract}
Foundation models have emerged as robust models with label efficiency in diverse domains. In medical imaging, these models contribute to the advancement of medical diagnoses due to the difficulty in obtaining labeled data. However, it is unclear whether using a large amount of unlabeled data, biased by the presence of sensitive attributes during pre-training, influences the fairness of the model. This research examines the bias in the Foundation model (RetFound) when it is applied to fine-tune the Brazilian Multilabel Ophthalmological Dataset (BRSET), which has a different population than the pre-training dataset. The model evaluation, in comparison with supervised learning, shows that the Foundation Model has the potential to reduce the gap between the maximum AUC and minimum AUC evaluations across gender and age groups. However, in a data-efficient generalization, the model increases the bias when the data amount decreases.
These findings suggest that when deploying a Foundation Model in real-life scenarios with limited data, the possibility of fairness issues should be considered.
\end{abstract}    
\section{Introduction}
\label{sec:Introduction}
% Introduzir Foundation Models, tecnicas de self-supervised
% citar que esses modelos são treinados em grandes bases de dados sem rotulos
% citar que eles são robustos a dados não vistos
% tem a propriedade de label effiency.
% comparar que tanto generative e constrative possuem tais propriedade
% entretanto o uso de grandes datasets no pre-treinamento sem rotulos dificulta a avaliação de vieses nesses datasets e consequentemente nos modelos pre-treinados.
% Devido a desigualdade de acesso a saude de uma forma global e a calote de dados muitos datasets possuem vieses principalmente
% falar sobre a importancia etica de termos datasets
% como a população latina pode ser impactada por não possuir grandes datasets
% vamos investigar se tais modelos possuem vies quando feito um finetuning em um dataset com uma população diferente da do pre treinamento
% Além disso iremos avaliar como a redução dos dados na hora do fine-tuning afeta a justiça do modelos.

Obtaining labeled data in the medical imaging domain presents considerable challenges, underpinned by the intricate and demanding nature of the annotation process \cite{krause_grader_2018}. This difficulty arises primarily from the need for highly specialized knowledge and expertise from trained medical professionals, who must meticulously examine and label medical images. However, acquiring images is still relatively easier, especially in the field of retinal imaging, where standardized imaging techniques like color fundus photography (CFP) and optical coherence tomography (OCT) are widely used for the diagnosis and follow-up of ocular diseases \cite{noauthor_optical_nodate-1}. These methods enable the consistent and non-invasive capture of high-quality retinal images, facilitating a more straightforward collection process compared to other types of medical imaging.

Recently, the emergence of self-supervised learning techniques such as Masked Autoencoder (MAE) \cite{chen_simple_2020} and Simple Contrastive Learning of Representations (SimCLR) \cite{he_masked_2021} for generative and contrastive learning has enabled the use of large unlabeled datasets to train models with significant capacity. This new paradigm has facilitated the emergence of Foundation Models (FM), which are trained on extensive datasets and can subsequently be adapted to a wide range of downstream tasks. Recent works investigate the potential of Foundation Models to create more robust models and data-efficient generalization in medical imaging using both contrastive learning \cite{azizi_robust_2023} and generative learning \cite{zhou_foundation_2023}.

Investigating bias in Foundation Models within the medical domain is essential to prevent the perpetuation of healthcare disparities \cite{rajkomar_ensuring_2018}. These models, trained on large and diverse datasets, can inadvertently incorporate biases related to race, gender, or socio-economic status, leading to skewed medical diagnoses or treatments. Such biases risk undermining the accuracy and fairness of medical assessments for underrepresented groups, highlighting the need for systematic bias detection and correction in these models. Ensuring that Foundation Models operate equitably is vital to maintaining trust and integrity in medical imaging and healthcare, necessitating rigorous evaluation and mitigation of potential biases \cite{bommasani_opportunities_2022}.

The contributions are summarized as follows:

\begin{itemize}
    \item We analyze bias in a retinal imaging Foundation Model (RetFound) using generative learning and fine-tuning on a Brazilian dataset. Our findings show that self-supervised learning effectively reduces biases compared to supervised learning.
    \item We found an increase in the gap between the maximum AUC and the minimum AUC for age when RetFound is trained with fewer data, highlighting a trade-off between data-efficient generalization and group fairness.
\end{itemize}
\section{Related works}
\label{sec:formatting}

The FM models developed for the domain of medical imaging exhibit robust performance on out-of-distribution data. These results indicate that such models can be adapted to other hospitals with varying data distributions \cite{azizi_robust_2023, zhou_foundation_2023, chen_towards_2024}. However, these studies do not investigate whether FM contributes to fairness issues, such as the extent to which data reduction affects performance for the less represented population in the dataset.

Previous studies have highlighted racial and sex-related biases in downstream tasks when using FMs trained with contrastive learning \cite{glocker_risk_2023}. This suggests that these models might pose risks for clinical applications. The fairness benchmark for medical images evaluates several techniques for mitigating bias in pre-processing, in-processing, and post-processing \cite{zong_medfair_2023}. However, these studies do not explore self-supervised learning as a method to mitigate bias.

\section{Methods}

The objective of this study is to systematically assess the presence of biases in FM and evaluate their efficacy in bias mitigation relative to supervised learning models. For this purpose, a Brazilian dataset is employed for the downstream binary classification task in diabetic retinopathy (DR), as detailed in Section \ref{sec:study_sample}. An FM was utilized alongside the development of a model from scratch using supervised learning (Section \ref{sec:model}). Both models were assessed using standard metrics in fairness evaluation to facilitate a comparative analysis of the techniques (Section \ref{sec:experiments}).

\subsection{Dataset}
\label{sec:study_sample}
During the study, we used the Brazilian Multilabel Ophthalmological Dataset (BRSET) \cite{nakayama_brazilian_nodate}, an ophthalmological dataset consisting of 8,524 Brazilian patients with 16,266 images. This dataset includes sex, age, and nationality as protected attributes. However, since all patients have Brazilian nationality, this feature was discarded. The target variable is diabetic retinopathy, with only 6.4\% of patients diagnosed with the condition. To work with the dataset, we eliminated patients without age or gender information and balanced the classes to 50\% with diabetic retinopathy and 50\% without, resulting in 1,097 patients with 1,416 images.

\begin{figure}[ht]
    \centering
    \includegraphics[width=1\linewidth]{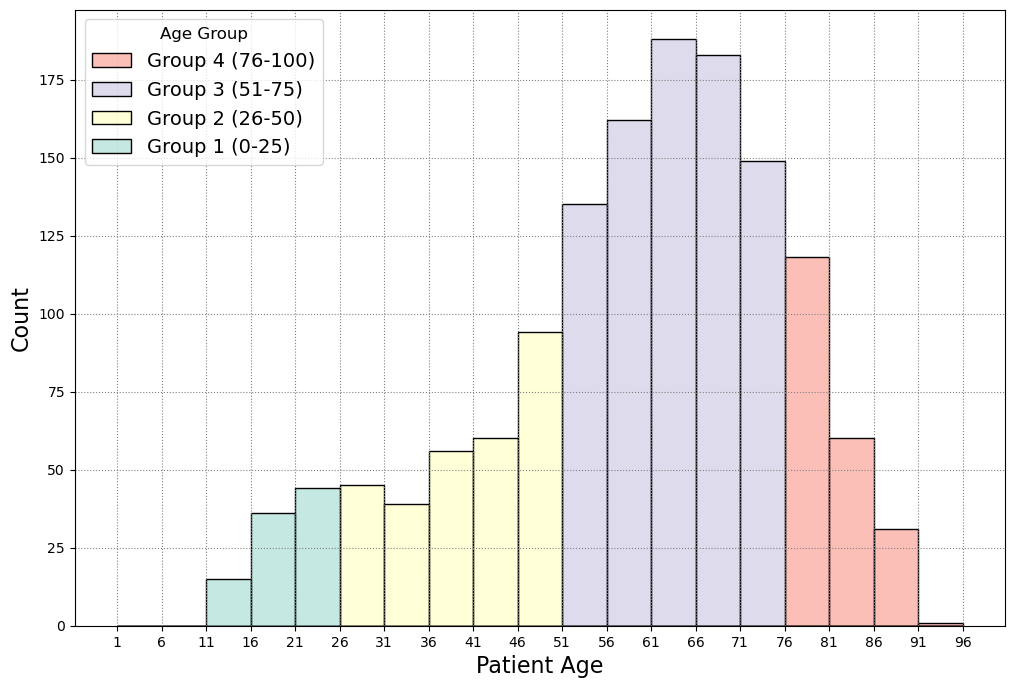}
    \caption{This histogram displays the age distribution from the pre-processed BRSET dataset, categorizing patients into four age groups: 0-25, 26-50, 51-75, and 76-100 years. This categorization facilitates the evaluation of the max-min metrics presented in \cref{sec:experiments}. }
    \label{fig:age_distribution}
\end{figure}

To conduct a structured analysis of age demographics, we segmented the population into four distinct age cohorts: 0-25 years, 26-50 years, 51-75 years, and 76-100 years. As shown in \cref{fig:age_distribution}, the main discrepancy in the dataset occurs in age, where the group aged 51 to 75 years has the largest number of samples with the diagnosis, comprising 817 patients or 57\% of the dataset. In terms of gender, the distribution is 55.72\% male and 44.28\% female, indicating another imbalance in the dataset.

To avoid data leakage, since the dataset consists of images of pairs of patients’ eyes, we separated the training and test datasets. We used images sized at $224 \times 224$ pixels and applied horizontal rotation as data augmentation, as rotating these images does not influence the model’s understanding of the dataset incorrectly. Additionally, we applied Color Jitter to simulate images with different lighting conditions and Random Erasing so that the model can handle capture errors, obstructions, or artifacts.

\subsection{Model}
\label{sec:model}
The current study uses the FM RetFound model \cite{zhou_foundation_2023} to gain insights into fairness problems in generative learning. This model is beneficial for study because it is pre-trained using MAE \cite{chen_simple_2020} successively on natural images (ImageNet-1k), followed by retinal images using 904,170 color fundus photographs (CFP), of which 90.2\% came from the Moorfields Diabetic Image Dataset (MEH-MIDAS) and 9.8\% from Kaggle EyePACS \cite{gulshan_development_2016}. These two datasets do not include images from South America, making the BRSET dataset an excellent resource for evaluating the model’s performance on an unfamiliar population.

For the downstream task, we employ a binary classification approach to detect diabetic retinopathy (absent DR and present DR, defined as International Diabetic Retinopathy Classification $\geq$ 1) \cite{wilkinson_proposed_2003}. Diabetic retinopathy is the leading cause of blindness worldwide and one of the most explored diseases in automated systems, with FDA-approved AI systems for screening referable cases. We fine-tune the ViT-L model, leveraging the pre-trained weights from RetFound \cite{zhou_foundation_2023}. Optimization is carried out using the Rectified Adam (RAdam) optimizer \cite{liu_variance_2021} with a learning rate of $10^{-4}$, a batch size of 16, and training over 50 epochs. The model weights that achieve the highest Area Under the Receiver Operating Characteristic (AUROC) on the validation set are retained as the checkpoint for subsequent internal and external evaluations. To identify if supervised learning is better than self-supervised learning in terms of fairness, we utilize the same architecture for comparison, training a ViT-L from scratch for 50 epochs for binary classification using the BRSET subset.

\subsection{Experiments}

For the assessment of bias in the Foundation Model (FM) and comparison, we utilize the same three aspects established in the fairness benchmark framework \cite{zong_medfair_2023} to enable comparison with other techniques and image domains used in the study. The first aspect is the utility of the model, measured by the Area Under the Receiver Operating Characteristic Curve (AUC) across all samples. The second aspect is group fairness, defined as the AUC gap between the subgroups with the maximum AUC and the minimum AUC. The third aspect is Max-Min fairness, which refers to the AUC of the group in the worst-case scenario.

The models are evaluated using 60\% of the data for training, 10\% for validation, and 30\% for testing. To ascertain the model’s fairness regarding data-efficient generalization, the training data sample was exclusively reduced while maintaining constant proportions for validation and testing.
\label{sec:experiments}

\section{Results and discussion}

To address the research question of whether the FM exhibits bias concerning sensitive attributes, a comparative analysis was conducted focusing on gender and age as primary variables. The assessment utilized three distinct metrics: Utility, Group Fairness, and Max-Min Fairness.

The first set of analyses examined the utility of the model as evaluated. Table \ref{tab:auc_data_models} compares the utility between the Baseline model, developed from scratch, and the fine-tuned RetFound model on the BRSET dataset. This comparison is structured to assess data-efficient generalization across different data percentiles in the training process. Various studies have assessed the efficacy of FM in data-efficient generalization \cite{azizi_robust_2023, zhou_foundation_2023, chen_towards_2024}. In \cref{tab:auc_data_models}, there is a clear trend of data-efficient generalization in RetFound as well as the strong supervised baseline. With only 25\% of the data used for training, we achieve superior results compared to the baseline trained on 100\% of the data.

\begin{table}
  \centering
  \begin{tabular}{@{}lcc@{}}
    \toprule
    Data Quantity & RetFound & Baseline \\
    \midrule
    100\% & 91.7 & 65.3 \\
    75\% & 86.5 & 63.1 \\
    50\% & 81.3 & 62.8 \\
    25\% & 68.9 & 60.7 \\
    10\% & 53.6 & 56.7 \\
    \bottomrule
  \end{tabular}
  \caption{AUC Results for Different Data Quantities in RetFound and Baseline Models. The table demonstrates the performance of each model across various levels of data availability.}
  \label{tab:auc_data_models}
\end{table}

\begin{figure*}[!htbp]
    \centering
    \includegraphics[width=0.8\linewidth]{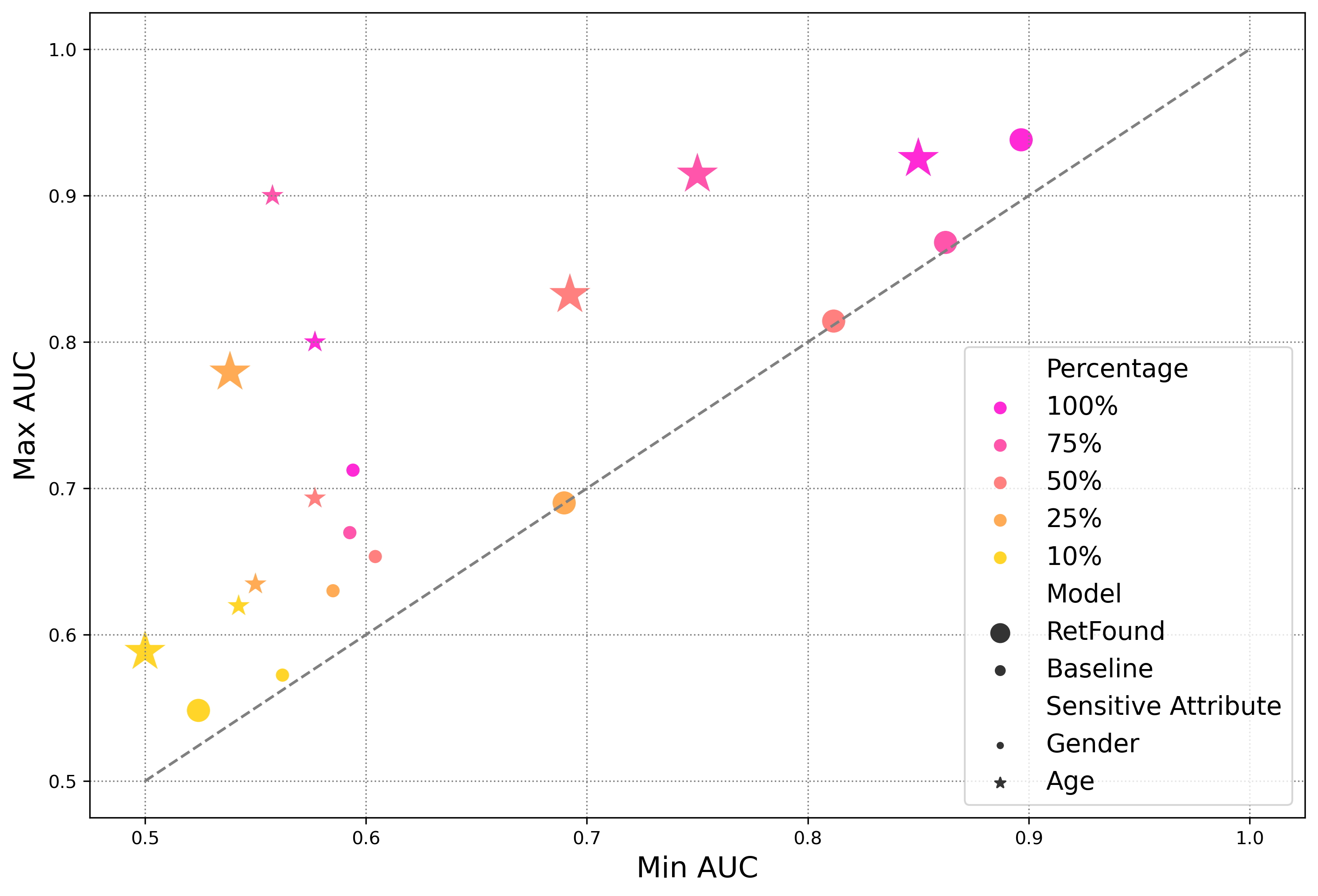}
    \caption{The scatter plot delineates the relationship between minimum and maximum AUC values for two models: ViT-L with RetFound weights and a baseline model developed from scratch. Point size differentiates the models, while colors and distinct markers denote data percentages and sensitive attributes, respectively. A reference grey line enables performance comparison, succinctly encapsulating AUC dynamics and attribute interactions.}
    \label{fig:max-min}
\end{figure*}

Further analysis of group fairness shows that the FM models do not exhibit significant disparities regarding gender. As shown in \cref{fig:max-min}, the points for RetFound and gender are close to the identity curve (grey line). However, significant gender differences were found for the baseline model. Primarily, as the size of the dataset increases, we observe larger disparities. For instance, employing merely 10\% of the dataset yields a satisfactory outcome for group fairness, with a disparity of 1\%. However, as the training data volume increases to 50\%, the disparity widens to 4.9\%, and further amplifies to 11.8\% when the dataset is fully utilized (100\%). This trend underscores the impact of dataset size on the manifestation of fairness discrepancies within the models. It is expected that with a larger volume of data, there exists an increased probability of identifying noteworthy disparities among various subgroups of the sensitive attribute.

Interestingly, employing the entire dataset (100\%) for fine-tuning the RetFound model results in a group fairness disparity of 4.1\%. However, this disparity diminishes significantly when training involves only 75\% of the data, where group fairness decreases to 0.6\%. Moreover, with further reductions in data usage to 50\%, 25\%, and down to 10\%, the fairness disparity continues decreasing, registering at 0.2\%, 0.03\%, and then increasing slightly to 2.4\%, respectively. These findings suggest that using smaller datasets for fine-tuning can sometimes lead to more equitable results, excluding the results with 10\% training data.

Considering the experimental evidence on the age-sensitive attribute, it is observed that this protected characteristic exhibits more substantial disparities compared to gender, as shown in \cref{fig:max-min}, where larger differences between the distributions of each group are evident. The baseline model demonstrates pronounced disparities across all datasets, with the most substantial discrepancy occurring with 75\% of the training data, reaching a disparity of 34.2\%. Utilizing the entirety of the dataset, the observed disparity stands at 22.3\%. A reduction in this gap is evident when smaller datasets are employed; when the dataset is limited to 25\% and 10\%, the disparities diminish to 8.46\% and 7.73\%, respectively. These findings suggest that the supervised approach benefits from using less data in terms of group fairness; however, the utility is worse, as can be seen from \cref{tab:auc_data_models}.

This outcome is contrary to that of RetFound, where an increase in data volume correlates with a reduction in the disparity concerning age, as delineated in \cref{fig:max-min}. When using 100\% of the data, the disparity narrows to 7.5\%, marking the minimum gap observed in the age attribute. Conversely, diminishing the dataset size leads to an incremental rise in the gap, escalating to 16.4\%, 13.9\%, and 24\% for 75\%, 50\%, and 25\% of the data, respectively. A notable improvement is observed only when utilizing 10\% of the data, where the gap marginally increases to 8.8\%. These differences are also expected, as the data is randomly sampled. When the percentage decreases, classes with low data volume may experience low representation in the new dataset and may even disappear when considering only 10\% of the data.

The results indicate that the self-supervised learning strategy employed in the RetFound model may offer advantages in reducing bias within downstream tasks, relative to the supervised baseline methodology. Specifically, an increase in data volume correlates with a decrease in age-related bias in the RetFound model. In contrast, an inverse relationship is observed for the baseline model, where increased data volume exacerbates bias.

% \maira{conclusão?}The data-efficient generalization is very important in the medical domain, that have difficulties obtaining labeled data \cite{azizi_robust_2023}, however, the findings in group fairness in this work suggest that the RetFound has large gaps when utilizing small datasets, showing the need for techniques to mitigate the bias, especially in small datasets. 

The third set of analyses examined Max-Min fairness. The results for the Min AUC of the supervised learning baseline do not exceed 0.6 for age and gender, even with increased data. However, referring to \cref{tab:auc_data_models}, the utility increases, primarily due to the Max AUC. As shown in \cref{fig:max-min}, the Max AUC significantly improves with an increase in data, reaching 0.9. There was a significant improvement in Min AUC when using a self-supervised approach with RetFound. The Min AUC reached 89.6\% for gender and 85\% for age.

\section{Conclusion}

This work evaluates the bias in a retinal images Foundation Model on a Latin American dataset for CFP. Observing the results for sensitive attributes, gender, and age, and comparing the results with a model using supervised learning, we found that the Foundation Model has less bias than supervised learning. However, when we evaluate the Foundation Model using less data, the bias increases specifically for age, but this same result does not occur for gender. The results of this study open the discussion on the trade-off between utility and group fairness in data-efficient generalization. Data-efficient generalization is very important in the medical domain, which faces difficulties in obtaining labeled data, and finding techniques to mitigate this gap in group fairness is essential for deploying equitable models in production. Future studies should investigate other databases to verify if the results are specific to this database or are generalizable to others. For future research, we will propose a reduction in the percentage of training data considering stratified sampling by subgroups of each sensitive attribute. This approach will maintain the distributions of each class, allowing us to visualize the differences in utility and fairness compared to reduction by random resampling.
{
    \small
    \bibliographystyle{ieeenat_fullname}
    \bibliography{main}
}


\end{document}